# Super-resolution of spatiotemporal event-stream image captured by the asynchronous temporal contrast vision sensor

**Hongmin Li[1,2], Guoqi Li[1,2*] , Hanchao Liu[1], Luping Shi[1,2*],**

[1]Center for Brain-Inspired Computing Research (CBICR), Department of Precision Instrument, Tsinghua University, Beijing 100084, China

[2]Beijing Innovation Center for Future Chip, Beijing 100084, China

## Abstract

Super-resolution (SR) is a useful technology to generate a high-resolution (HR) visual output from the low-resolution (LR) visual inputs overcoming the physical limitations of the cameras. However, SR has not been applied to enhance the resolution of spatiotemporal event-stream images captured by the frame-free dynamic vision sensors (DVSs). SR of event-stream image is fundamentally different from existing frame-based schemes since basically each pixel value of DVS images is an event sequence. In this work, a two-stage scheme is proposed to solve the SR problem of the spatiotemporal event-stream image. We use a nonhomogeneous Poisson point process to model the event sequence, and sample the events of each pixel by simulating a nonhomogeneous Poisson process according to the specified event number and rate function. Firstly, the event number of each pixel of the HR DVS image is determined with a sparse signal representation based method to obtain the HR event-count map from that of the LR DVS recording. The rate function over time line of the point process of each HR pixel is computed using a spatiotemporal filter on the corresponding LR neighbor pixels. Secondly, the event sequence of each new pixel is generated with a thinning based event sampling algorithm. Two metrics are proposed to assess the event-stream SR results. The proposed method is demonstrated through obtaining HR event-stream images from a series of DVS recordings with the proposed method. Results show that the upscaled HR event streams has perceptually higher spatial texture detail than the LR DVS images. Besides, the temporal properties of the upscaled HR event streams match that of the original input very well. This work enables many potential applications of event-based vision.

## 1    Introduction

DVS (Lichtsteiner, Posch, and Posch 2008, Serrano-Gotarredona and Linares-Barranco 2013, Yang, Liu, and Delbruck 2015) is a kind of frame-free dynamic vision sensors whose principles of operation are based on abstractions of the functioning of biological retinas. Because of large pixels

---

with relatively small fill factors of the event-based sensors (Lichtsteiner, Posch, and Posch 2008), the DVS image has low spatial resolution, which lead the spatiotemporal texture to be indistinct. The event-stream image super-resolution (SR) refers to the task of obtaining the high-resolution (HR) enlargement of low-resolution (LR) DVS recordings. Different from synchronous frame-based output of traditional cameras, the DVS pixel asynchronously and independently encodes the local intensity contrast into precisely time-stamped events. The DVS images are in the format of spatiotemporal address-event representation (AER) event streams (Li et al. 2017, Orchard et al. 2015, Serrano-Gotarredona and Linares-Barranco 2015, Tan, Lallee, and Orchard 2015). The DVS image SR problem arises in various real-world applications. A common application occurs when we want to enlarge a LR DVS recording which are usually with $128 \times 128$ resolution due to the limited physical size of the sensor for display purpose. For example, when we click a DVS recording shown in web pages, we often want to see the corresponding enlarged HR version. Another application is found when we want to magnify the tiny spatiotemporal event-stream objects in the pattern recognition or visual detection task. When a DVS is used in a microscopes to capture the movement of tiny creatures, or physical particles, the event-based image is supposed to be enlarged to recover the texture detail. For the traditional computer vision, many SR works have been proposed (Glasner, Bagon, and Irani 2009, Hou and Andrews 1978, Xu, Qi, and Chang 2014, Keys 1981, Yang et al. 2008, Wang et al. 2015), and achieved good performance. But almost all the traditional methods are to process two-dimensional images and cannot be directly applied to DVS recordings. Besides, many impressive works (Reinbacher, Graber, and Pock 2016, Kim et al. 2008, Bardow, Davison, and Leutenegger 2016) have tried to recover the HR 2-dimensional intensity field or the grayscale value of each pixel from DVS recordings. In (Kim et al. 2008), the authors demonstrated a high quality joint estimation of scene intensity and motion from pure event data with pure camera rotation in an otherwise static scene. In (Bardow, Davison, and Leutenegger 2016), authors proposed a method to simultaneously recover the motion field and brightness image from event streams. Most of the works aim to address the intensity image reconstruction problem based on the fast DVS sensing. However, to the best of our knowledge, there are still no methods available to obtain the HR spatiotemporal event stream from a LR DVS recording.

The key problem is how to generate the event sequence of each pixel of the HR event-stream image. In this work, we model the event sequence of each pixel with a Poisson point process(Eden et al. 2004, Truccolo et al. 2005), and sample the events by simulating a nonhomogeneous Poisson point process with the specified event number and rate function. The event of DVS has the property of all-or-none. Many kinds of all-or-none event process including the occurrences of major freezes in Lake Constance (Steinijans 1976), neural spike train data analysis(Eden et al. 2004, Truccolo et al. 2005), and transaction processing in a data base management system(Lewis and Shedler 1976), can be modeled using point process. Besides, many works (Perkel, Gerstein, and Moore 1967a, b) have discussed the relationship between spike trains and point process. In a Poisson point process, the event numbers in any finite set of non-overlapping intervals are mutually independent random variables of Poisson distribution. Generally, the general Poisson process is defined as a non-decreasing right-continuous function $\Lambda(t)$ which is bounded in any finite interval. Let the counting process $N(t)$ denotes the number of events that have occurred before $t$. The $\Lambda(t)$ function is called the integrated rate function, that is, for $t \geqslant 0$, $\Lambda(t) - \Lambda(0) = E[N(t)]$. The right derivative $\lambda(t)$ with respect to $\Lambda(t)$ is the intensity function or rate function which represents the instantaneous arrival rate at time $t$.

Firstly, the event number of each pixel of the HR event-stream image should be determined. We denote the event number of each pixel of the event-stream image with an event-count map of which each entry represents the number of events. The event-count map of the LR DVS input can be generated by accumulating the events of each pixel in a time window. We aim to generate the HR event-count map from the LR event-count map with a resolution enhancement method. Many efficient methods including interpolation based methods and example-based methods, have been proposed to achieve the resolution enhancement for the traditional matrix-based image (Asamwar, Bhurchandi, and Gandhi 2010, Dong et al. 2014, Glasner, Bagon, and Irani 2009, Hou and Andrews 1978, Keys 1981, Wang et al. 2015, Xu, Qi, and Chang 2014, Yang et al. 2008, Zhang and XiaolinWu 2008, Zhou, Dong, and Shen 2012).

Secondly, two main methods exist to simulate a nonhomogeneous Poisson process. One is the time-rescaling theorem based method (Gerhard and Gerstner 2010, Haslinger, Pipa, and Brown 2010, Brown et al. 2002) which assumes that any nonhomogeneous Poisson process can be transformed into a homogeneous Poisson process. For many Poisson processes, the inversion of $\Lambda(t)$ function is complicated and needs to be computed numerically. Then this time-rescaling theorem based method may be far less efficient. Different from the time-rescaling method, the other method named "thinning" (Lewis and Shedler 1979, Ogata 1981) is not only conceptually simple, but also computationally simple and relatively efficient, without the need for numerical integration. In thinning method, a point process is sampled by accepting or rejecting the events originally generated from a homogeneous Poisson process because the homogeneous Poisson processes can be generated efficiently in a straight-forward fashion. In this work, the events of each new pixel are sampled based on thinning algorithm.

In this work, a two-stage scheme is proposed to implement SR of spatiotemporal event streams. In the first stage, the event number and rate function of each pixel of the expected HR event-stream image are generated. In the second stage, events of each pixel of the HR event-stream image are generated by simulating a nonhomogeneous Poisson process according to the specified event number and rate function. Several experiments are conducted to demonstrate the effectiveness of the proposed method, including obtaining HR event streams from a series of DVS recordings. We displayed the results with the reconstructed frames by integrating the ON and OFF events and the temporal properties of total events using the firing rate of all the events. The HR event-stream images show more texture detail visually compared to the LR DVS image. The total rate function curves of the upscaled HR and original LR event streams match very well.

This paper proceeds with an introduction to temporal contrast pixel of the DVS sensor in Section 2.1, before describing the general architecture of the proposed method in Section 2.2. In Section 3, we present the experiments and results before wrapping up with conclusion in Section 4.

## 2    Materials and method

### 2.1    Event-based Vision Sensor

The DVS sensor used in this work is a 128×128 pixel device (Lichtsteiner, Posch, and Posch 2008). As shown in **FIGURE** 1, each pixel independently and continuously measures local intensity change and generates an event whenever the log intensity change exceed a pre-defined threshold since the last emitted event. An ON or OFF event is generated depending on whether the log light intensity increased or decreased. These events appear at the output of the sensor as an asynchronous stream of digital pixel addresses. The temporal information of the dynamic scene is encoded into event sequence with sub-millisecond timing precision. In addition, the timing of events have a temporal resolution of 1 μs with the equivalent frame rate of typically several kHz.

Each event is defined as a quadruplet $e(u, t, p) = [u, t, p]^T$, where $u = [x, y]^T$ is the pixel position, $t$ is the timestamp of the event and $p$ is its polarity that can be +1 (ON) or -1 (OFF). The event rate is proportional to the change of log light intensity as follows

$$f(t) = \frac{1}{\theta} \frac{d}{dt} \ln(I(t)) \qquad (1)$$

where $I(t)$ denotes the light intensity on the pixel, and $\theta$ denotes the pre-defined threshold contrast.

Generally, the event rate is a time-varying function due to the local intensity changes of the visual scene. Two kinds of important information are contained in the DVS recording. One is the event number which represents how many times the temporal contrast exceeds the threshold. The other is the event rate or rate function which reflects the temporal property of the dynamic scene.

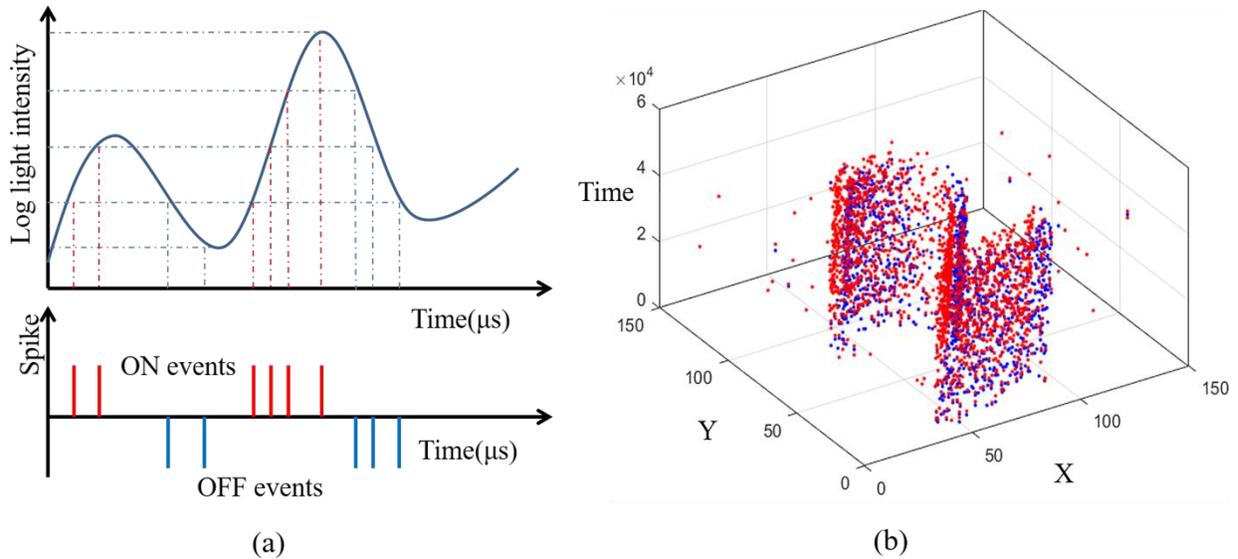

(a)                                                                (b)

**FIGURE 1** | Visual sensing of event-based vision sensor. (a) Principle of DVS temporal contrast pixel. An event is generated when the intensity change exceeds a pre-defined threshold contrast. The polarity is ON (red event) if the intensity increases, otherwise OFF (blue event). (b) Event streams of a digit 2 captured by a DVS sensor. Red point is ON event, and blue point is OFF event.

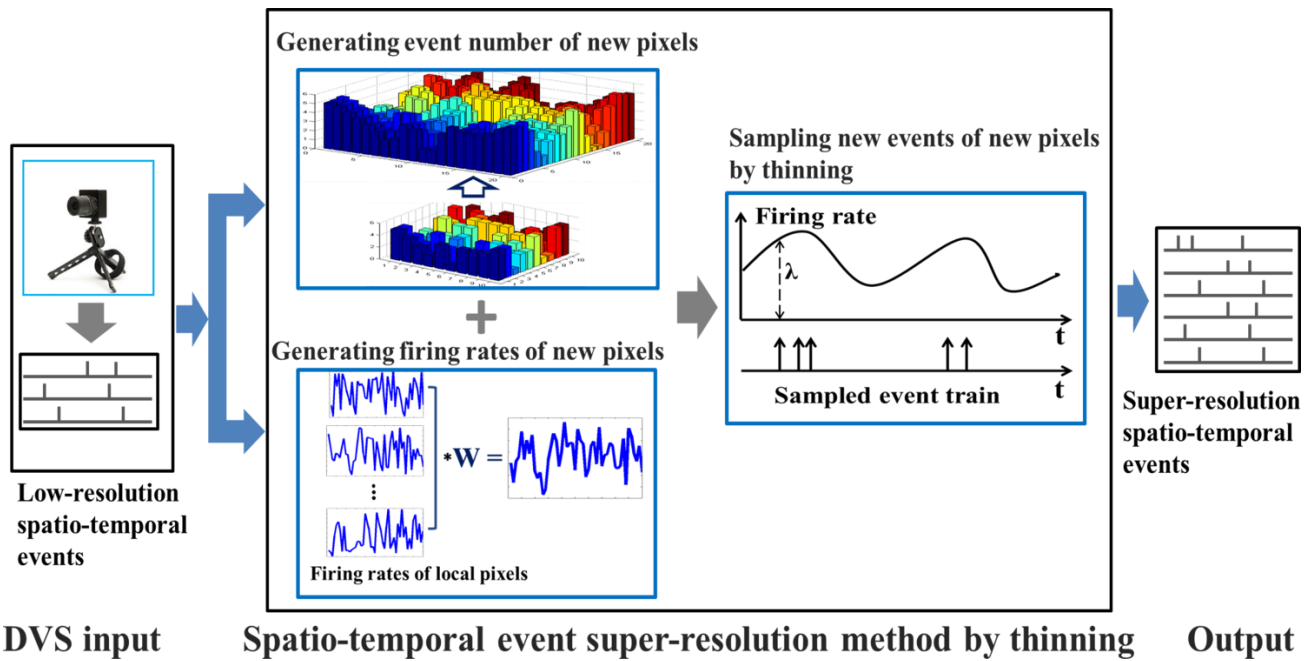

**DVS input        Spatio-temporal event super-resolution method by thinning        Output**

**FIGURE 2** | General architecture of the proposed two-stage method. Output of DVS sensor is usually spatiotemporal event streams. In the first stage, an event-count map and rate function of each pixel of the expected HR event stream are generated. In the second stage, events of each pixel of the HR recording is sampled

by simulating a nonhomogeneous Poisson process depending on the previous obtained event-count map and rate functions.

## 2.2   General Architecture

The general architecture of the proposed two-stage scheme is shown in **FIGURE** 2. Two problems should addressed in this scheme as follows: 1) how to generate the event number and rate function of each new pixel in the HR event-based image, 2) how to further generate the corresponding temporal events of each HR pixel. As the event number of each pixel is contained in the event-count map, we should obtain the event-count map of the HR event-stream image by upsampling that of the LR DVS input. The event-count map of the LR input can be generated by accumulating the events of each pixel in a window. The rate function of each HR pixel is obtained using a local spatiotemporal filter on the local pixels near the corresponding pixel of the LR DVS input. Then new events of each HR pixel are sampled by simulating a one-dimensional nonhomogeneous Poisson point processes according to the obtained event-count map and rate functions.

### 2.2.1 Simulation of Nonhomogeneous Poisson Process

We use a one-dimensional nonhomogeneous Poisson process to model the event sequence of each pixel. In this work, the simple and efficient thinning method (Lewis and Shedler 1979) is used to simulate a one-dimensional nonhomogeneous Poisson process. Given the rate function $\lambda(t)$ in a fixed time window, events are sampled by sampling a sequence of events of an easily simulated Poisson process with rate function $\lambda*(t)$. Considering a point process with rate function $\lambda*(t)$ in the time interval $(0, T)$, let $S_1^*, S_2^*, \cdots, S_{N^*(T)}^*$ be the events of the point process in the interval $(0, T]$. By thinning, a nonhomogeneous Poisson process with rate function $\lambda(t)$ ($\lambda(t) \leq \lambda*(t)$) in the interval $(0, T]$ can be obtained by deleting the event $S_i^*$ (for $i = 1, 2, 3, \ldots, n$) with probability $1 - \lambda(S_i^*) / \lambda*(S_i^*)$. The efficiency of the thinning algorithm is measured by the number of events deleted or accepted, and is proportional to $(\Lambda(t0) - \Lambda(0)) / (\Lambda*(t0) - \Lambda*(0))$ which is the ratio of the areas between 0 and t0 under $\lambda(t)$ and $\lambda*(t)$. Then the rate function $\lambda*(t)$ should be as close as possible to $\lambda(t)$. In the more simple and more efficient form of thinning method, the nonhomogeneous Poisson process $N*(t) = \max\{n \geq 0: S_n^* \leq t\}$ can be replaced by a homogenous Poisson process with $\lambda*(t) = \lambda*$. **FIGURE** 3 shows that events of the expected nonhomogeneous Poisson process are generated by thinning a homogeneous Poisson process with a constant intensity $\lambda* > \lambda(t)$. In this paper, an event sequence with the specified event number and rate function is simulated based on the ***Event Sampling Algorithm***. Then to sample the events of the HR event-stream output, we should specify the event number and rate function of the point process of each pixel.

***Event Sampling Algorithm***: Generate an event sequence with the expected event number.
   ***Input***: the time interval $(0, T]$, the expected number of events to be sampled $N$, the expected rate function $\lambda(t)$.
   1.   Set $j = 1$.
   2.   Generate $N*$ events of a Poisson process with rate function $\lambda*$ in the time interval $(0, T)$.
   3.   Denote the events by $S_1^*, S_2^*, \cdots, S_{N^*}^*$, Set $i = 1$.
   4.   While $i \leq N*$
      ◎   Generate a random number $U_i$ independently from a uniform distribution between 0 and 1,
      ◎   If $U_i \leq \lambda(S_i^*) / \lambda*$, $S_j = S_i^*$, and then set $j$ equal to $j+1$.
      ◎   If $j > N$, stop.

◎ Set $i$ equal to $i$+1

5. If $j \leq N$, go to step 2. Otherwise stop.

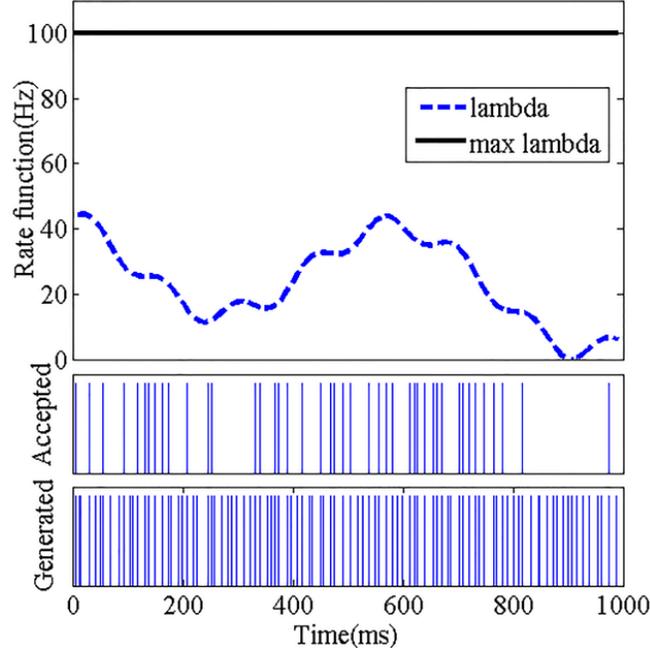

**FIGURE 3 |** Simulation of a Poisson process by thinning. Top: The solid line is the rate function of a homogeneous Poisson process. The dash line is the rate function of the nonhomogeneous Poisson process to be simulated. Bottom: the generated events are from the homogeneous Poisson, and the accepted events are generated by simulating the nonhomogeneous Poisson process by thinning.

### 2.2.2 Event Number

The event-count map of LR DVS input can be easily obtained by counting the number of events of each pixel. To generate the event-count map of the HR output from a LR input, we use a resolution enhancement based method. The simplest resolution enhancement method is interpolation. However, many example-based methods (Dong et al. 2014, Yang et al. 2008) have achieved better performance. In this work, we generate the HR event-count map using a sparse signal representation (SSR) based method according to (Yang et al. 2008). The HR event map patch is inferred from each LR event map patch of the input. In this patch-based model, we have two coupled dictionaries $D_l$ and $D_h$: $D_h$ consists of HR patches and $D_l$ consists of corresponding LR patches. The HR patch is recovered from the $D_h$ directly using the sparse representation of the corresponding LR patch in terms of $D_l$. In our method, the two dictionaries are generated by randomly sampling raw patches from the training event-count maps of a set of DVS recordings. These simply prepared dictionaries have been demonstrated effective to generate high-quality reconstructions (Yang et al. 2008). The optimization problem of the method is formulated as follows:

$$\min \quad \|\alpha\|_1$$
$$s.t. \quad \|FD_l\alpha - Fy\|_2^2 \leq \varepsilon_1 \qquad (2),$$
$$\|PD_h\alpha - w\|_2^2 \leq \varepsilon_2$$

where $F$ is a feature extraction operator, $\alpha$ is the coefficients with very few nonzero entries, $D_l$ and $D_h$ are two coupled overcomplete dictionary, $D_l$ is for the LR patches and $D_h$ is for the HR patches, $y$ is the patch of the inputted LR event-count map, $P$ is used to extract the overlap region between current patch and previously reconstructed HR event-count map, $w$ contains the values of the previously reconstructed HR event-count map on the overlap. We reconstruct the HR event number patches using the sparse representations learned from the LR event number patches. As such, the HR event number patch is reconstructed as $x = D_h \alpha^*$ after the optimal solution $\alpha^*$ to (2) is gained. The training set of event-count maps is taken from a series of recorded DVS recordings. Finally, the overlapping reconstructed patches are aggregated to produce the HR event-count map.

### 2.2.3 Rate Function

The rate function of the point process of each pixel is a one-dimensional function across the time line. For the original LR input, the rate function of each pixel is represented as the peristimulus time histogram (PSTH). This histogram is constructed by dividing the time window into successive non-overlapped $N$ bins of size $\Delta t$, and then counting the number of events $k_i$ in the bin $i$. The rate function is calculated by normalizing the histogram with the maximum of all the bins. We assume that the rate function of each new pixel in the expected HR event streams is determined by the local neighbor pixels near the corresponding position of the LR DVS input. Motivated by the local image smoothing (Sonka, Hlavac, and Roger 2014) in image processing, a local spatiotemporal filter is used to generate the rate function of the new pixel from the rate functions of LR DVS recording. The filter can be represented as a tensor $\boldsymbol{\kappa(I_x, I_y, I_t)}$ which is a 3-dimensional spatiotemporal kernel. The rate function of each new pixel can be calculated as follows,

$$\lambda_H(x, y, t) = \sum_{i_x} \sum_{i_y} \int_{i_t} \kappa(i_x, i_y, i_t) \bullet \lambda_L(x + i_x, y + i_y, t + i_t) dt \quad (3),$$

where $\lambda_H$, $\lambda_L$ denotes the rate function of a pixel in the expected HR event-stream image and LR DVS input, respectively. The filter is discrete on the time coordinate due to that we divide the time window into a series of bins. Then the rate function can be modified as follows,

$$\lambda_H(x, y, t) = \sum_{i_x} \sum_{i_y} \sum_{i_t} \kappa(i_x, i_y, i_t) \bullet \lambda_L(x + i_x, y + i_y, t + i_t) dt \quad (4)$$

We take a simple form by set $\boldsymbol{I_t}$ to be 0, which implies that we ignore the time effect. Then in each time bin, the filter $\boldsymbol{\kappa(I_x, I_y, I_t)}$ is reduced to a 2-dimensional convolution kernel $\boldsymbol{\kappa(I_x, I_y)}$ similar to that in traditional image processing, and can be written as a matrix. The sum of all the entries of the matrix is 1. Here we use 9 local neighbor pixels to construct the rate function, and then the filter is a matrix of $3 \times 3$. If every entry except the center one of the matrix is 0, the filter is a nearest neighbor interpolation to generate the rate function of the new pixel. In this work, the filter we used is represented as follows,

$$W = \frac{1}{16} \times \begin{bmatrix} 0 & 1 & 0 \\ 1 & 12 & 1 \\ 0 & 1 & 0 \end{bmatrix}.$$

### 2.2.4 Super-Resolution of Event Streams

The proposed SR method will not process each event because a single event has no visual information, while it is a group of events that form the spatiotemporal shape of the object. Then in the method, a set of events in a time window is magnified for super resolution. The firing rate function represents the dynamic temporal property of the event cluster. For a DVS image of long-term dynamic process, the spatiotemporal event stream can be divided into multiple DVS bins. The proposed method is applied to upsample the event streams in each time window and generate the corresponding HR spatiotemporal event stream. And then the upscaled HR event stream in each time window is stitched together on the timeline. The positive and negative events are sampled separately and merged according to the timing. The proposed two-stage event streams SR method is as follows.

**Algorithm**: SR method of spatiotemporal event streams.
   **Input**: A LR DVS recording of event streams $ES^L$, and the magnification $\alpha$.
   **Output**: The upscaled HR spatiotemporal event streams $ES^H$.
   **Stage 1**: 1) Generate the event-count map $ENM^L$ of $ES^L$ by counting the event number of each pixel, and the rate function of each pixel of $ES^L$.
   2) Generate the HR firing rate frame $ENM^H$ based on resolution enhancement method from $ENM^L$.
   3) Generate the rate function $\lambda^H_{i,j}(t)$ for each pixel $(i, j)$ of the upscaled HR event streams $ES^H$ using a spatiotemporal filter $\kappa(x, y, t)$.

**Stage 2**: For each pixel $(i, j)$ of the expected HR event stream, sample an event sequence $e_{i,j}$ with the number of events specified in $ENM^H$ and the corresponding rate function $\lambda^H_{i,j}(t)$ according to the **Event Sampling Algorithm**.

## 3    Experiments and Results

We evaluate the proposed method by obtaining HR event-stream images from a variety of LR DVS recordings which can be accessed online[2]. **FIGURE 4** shows the reconstructed frames of all the used event-stream recordings captured by a DVS sensor for SR experiments. Nine event-stream recordings are used for evaluating the effectiveness of the proposed method, including the event streams of a banana, a hand, a digit 5, a wooden fish, a cup, a digit 2, a scene with five digits, a digit 3, and a fork. In the SR reconstruction experiment, the original image is assumed as the groundtruth. The DVS input is firstly downsampled to a LR version and then upsampled to a HR version. In the magnification experiment, the original LR DVS input is magnified to a HR event-stream image with the same temporal dynamics according to a specified factor. SR results is assessed by the spatial texture detail displayed as a reconstructed frame and the temporal dynamics of the event streams. The total number of events in HR recording is also increased by the same magnification factor. The reconstructed frame is constructed by integrating the ON and OFF events of each pixel and displayed as a grey-scale map. In order to compare the spatial texture details, the reconstructed frames of the outputted HR event-stream image and LR DVS input are displayed in an identical size.

The temporal dynamic of an event sequence can be visualized as a PSTH. In this work, we construct the PSTH of an event sequence with a time bin of 100 us. We introduce an evaluation criterion by comparing the time properties of each pixel between the HR event streams and the groundtruth. The criterion is measured by the root mean squared error (RMSE) of the PSTH of each pixel as,

$$RMSE = \sqrt{\frac{1}{W \bullet H} \sum_{i,j} \frac{1}{T_2 - T_1} \int_{T_1}^{T_2} \left[ f_h^{i,j}(t) - f_g^{i,j}(t) \right]^2 dt} \qquad (4),$$

where $f_h^{i,j}$ and $f_g^{i,j}$ denote the PSTH of the pixel $i,j$ of the HR event-stream image and the groundtruth, respectively. A small RMSE means a better SR result.

Besides, because of the absence of groundtruth in the magnification experiment, and then goodness of the fit of the firing rate of total events in the magnified HR image to that of the original DVS input is also assessed by the difference between the firing rate functions (DFRF). DFRF is calculated as the ratio of root mean square of the difference between the firing rate function $f_h(t)$ of HR event-based image and original firing rate $f_l(t)$ of the inputted LR DVS image to the mean of the original firing rate as follows,

$$DFRF = \frac{\sqrt{\dfrac{1}{T_2 - T_1} \int_{T_1}^{T_2} \left[ f_h(t) - f_l(t) \right]^2 dt}}{\dfrac{1}{T_2 - T_1} \int_{T_1}^{T_2} f_l(t) dt} \bullet 100\% \qquad (5).$$

For resolution enhancement of the event-count map, 66 event-count maps of 66 DVS recordings were used to train an over-complete dictionary for sparse representation. $3 \times 3$ LR patches with overlap of 1 pixel between adjacent patches are used, corresponding to $9 \times 9$ patches with overlap of 3 pixels for the patches of HR event-count maps. In this work, the length of bin in the rate function is set to 50 us is not specified.

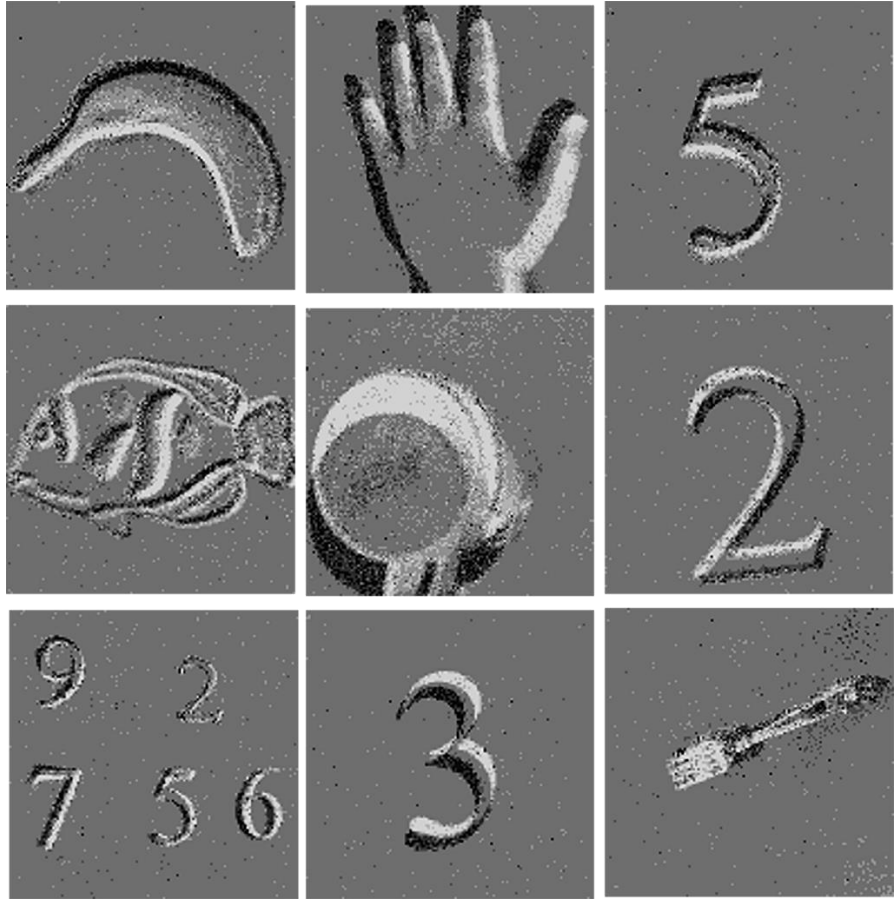

**FIGURE 4 |** Reconstructed frames of the nine event-stream DVS recordings on which to perform the SR experiments.

### 3.1 Super-Resolution Reconstruction

In this section, we evaluate the performance of the proposed method by SR reconstruction with several DVS recordings. We assume the original DVS recording is a HR version as the groundtruth. Firstly, the original DVS recording is downsampled by a factor of 2 to generate the LR event streams. Then the downsampled LR event streams are upscaled to the original size with the proposed method. **FIGURE 5** shows the results of the SR reconstruction of the five DVS recordings, i.e., the moving hand, digit 3, the scene of digits, cup, and wooden fish. All the reconstructed frames are displayed in the same spatial size. The texture details of the LR event streams, SR resulted event streams, and the groundtruth are compared. Results show that the reconstructed frames of the resulted SR event streams show higher texture and edge detail compared with that of the LR event streams which are visually smooth and blurry. The proposed method is demonstrated to be effective in recovering the high texture detail from the LR input.

### 3.2 Different Time Bins of Rate Function in Event Sampling

In the proposed method, a hyper-parameter, the length of the time bin $\Delta t$ of rate function $\lambda(t)$ in **_Event Sampling Algorithm_**, should be specified in the construction of the rate function of the Poisson point process of each pixel. This is because different length of time bins in the rate function will influence the precision of event sampling. In the experiment, four DVS recordings were used , i.e., the digit 3, digit 5, hand and wooden fish. Event streams of these recordings are all in an observation duration of 200 ms.  In the experiment, they are firstly downsampled to half of the spatial size and then upsampled with the proposed method under different time bins. A total of 27 different time bins are tested, i.e., 20 us, 40 us, 60 us, 80 us, 100 us, 200 us, 300 us, 400 us, 500 us, 600 us, 700 us, 800 us, 900 us, 1 ms, 2 ms, 3 ms, 4 ms, 5 ms, 6 ms, 7 ms, 8 ms, 9 ms, 10 ms, 20 ms,40 ms,80 ms,100 ms. are used. We also use the RMSE metric to test the influence of different time bins in PSTH rate function of event sampling. Four event-stream recordings are used in this test. **FIGURE 6** shows the RMSE results over different length of time bins. As the length of time bin increases, the RMSE increases. In other words, a smaller time bin lead to better SR result. Besides, the RMSE of the wooden fish is larger than the digits, because the wooden fish has more complicated textures than the scenes of digits. **FIGURE 7** shows the temporal dynamics of the LR input, groundtruth, and SR results under different time bins with some successive reconstructed frames in time line. The reconstructed frames is constructed every 40ms, and then 5 reconstructed frames over the time line are generated. Results show that as the length of time bin increases from 20 us to 1000 us, the SR event streams remains nearly the same temporal dynamics with the groundtruth. Then the proposed method is robust to the length of the bin. When the length is longer than 20 ms, the SR event streams become very different with the groundtruth. This is because the longer time bin in the rate function lead to low-precision event sampling in the thinning algorithm.

### 3.3 Robustness

In this section, we evaluate the robustness of the proposed method by repeating the SR reconstruction 10 times on each DVS recordings. All the nine DVS recordings are used for SR reconstruction. Each time, we compute the RMSE value and then get the maximum, minimum, mean and standard deviation of the 10 tests. Results in **Table 1** shows that the RMSE measurements of 10 times remain in a relatively small range with a small standard deviation. The maximum and minimum of RMSE is very close. Results demonstrate the robustness of the proposed method.

### 3.4 Different Magnification factors

In this section, several DVS recordings of different objects were magnified by a larger factor to evaluate the performance of the proposed method. We use the reconstructed frames and PSTH of all the events to visualize the SR results. In order to compare the spatial texture detail of the reconstructed frames, the reconstructed frame of the LR DVS recording is stretched to the same size with the magnified HR event streams. **FIGURE 8** shows the SR results of a DVS recording of a banana captured by a moving DVS by a magnification factor of 3. The reconstructed frames and total rate function curves of both the upscaled HR and original LR event streams are shown in **FIGURE 8 (a)**. The reconstructed frame of LR DVS recording shows overly-smooth and thus have poor perceptual quality. The reconstructed frame of the HR event streams shows visually high texture detail. The total rate curves of both the HR and LR event streams matches well. Besides, we also displayed the spatiotemporal events in a three-dimensional coordinate space as shown in **FIGURE 8 (b).** The spatiotemporal distribution of the HR events is visually consistent with that of the LR recording. Furthermore, we also evaluate a larger magnification factor of 4 for a DVS recording of simple texture. A digit 3 is captured by a moving DVS sensor. **FIGURE 9 (a)** shows that the upscaled DVS recording has visually clear textures. Besides, the two total rate function curves match very well with small DFRF error. From **FIGURE 9 (b),** we can see that the events of upscaled HR and original LR DVS recordings have almost the same spatiotemporal distribution. Then the proposed method is demonstrated effective in SR with larger magnification factor. **TABLE** 2 shows the DFRF results of SR with different magnification factors. All the recordings are magnified with two factors of $\times 3$ and $\times 4$. The DFRF with larger magnification factor lead to larger error between the rate functions of total events in the SR magnified event streams and the original input. Even so, the DFRF results are relatively small, which demonstrate the effectiveness of the proposed method with larger magnification factor.

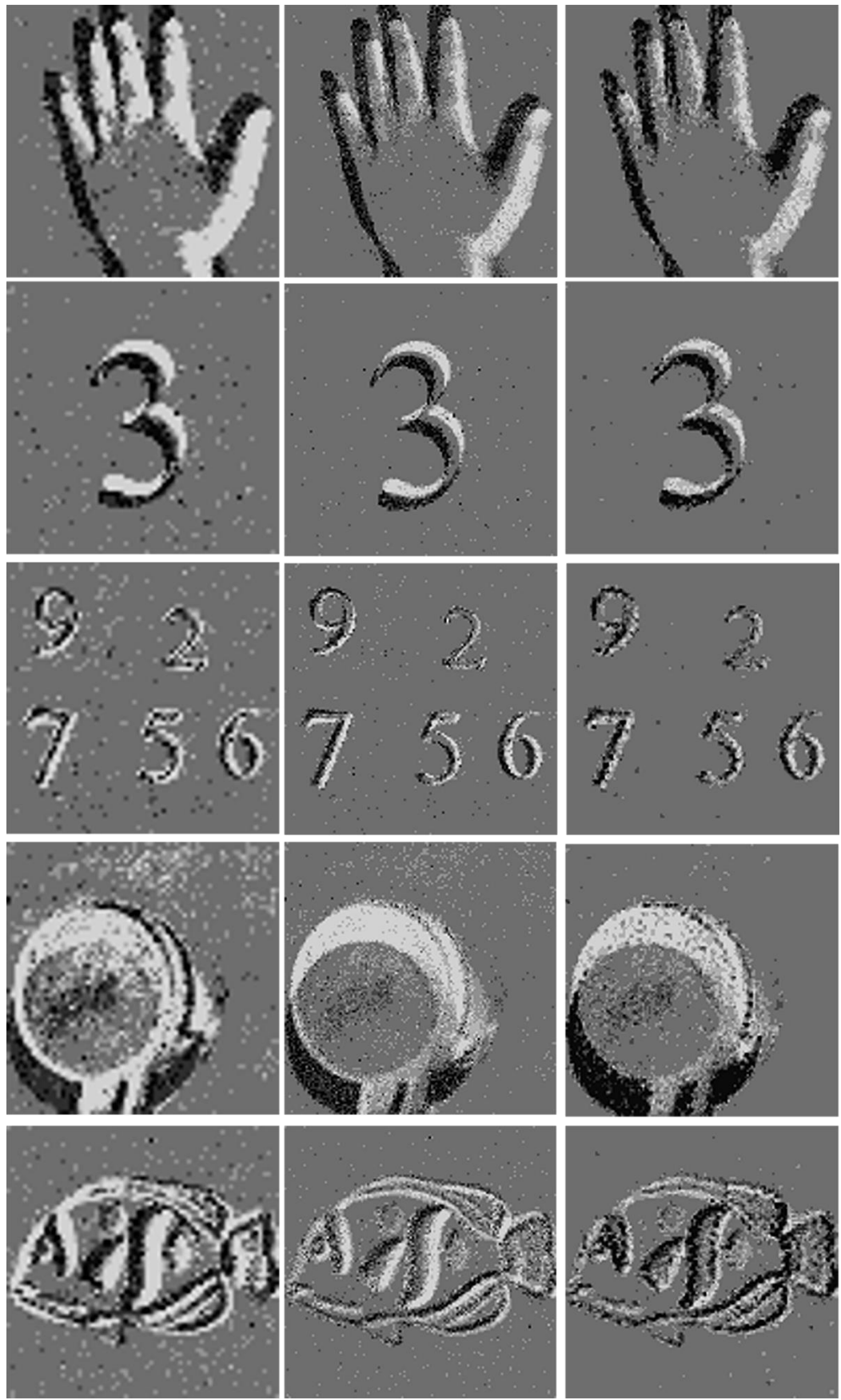

**FIGURE 5|** SR reconstruction of five DVS recordings with a magnification factor of 2. From left to right are reconstructed frames of downsampled LR DVS recordings, groundtruth, upsampled HR event-stream image.

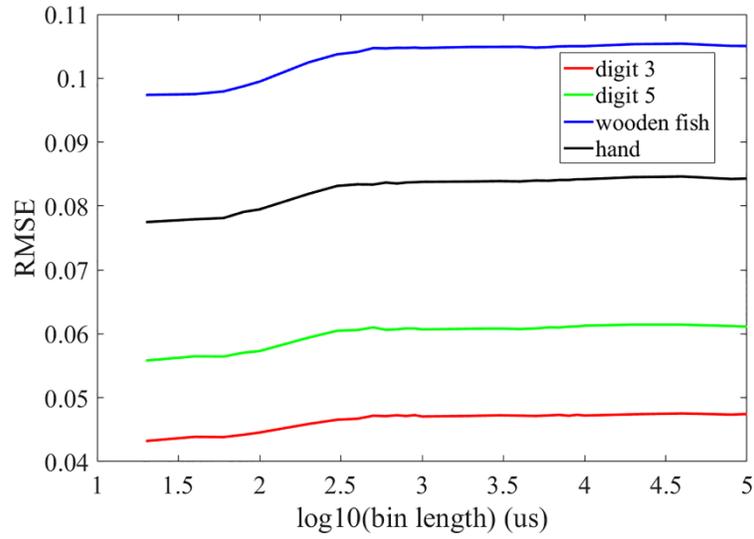

**FIGURE 6|** RMSE between the SR result and groundtruth over different length of the bins in the rate function of point process.

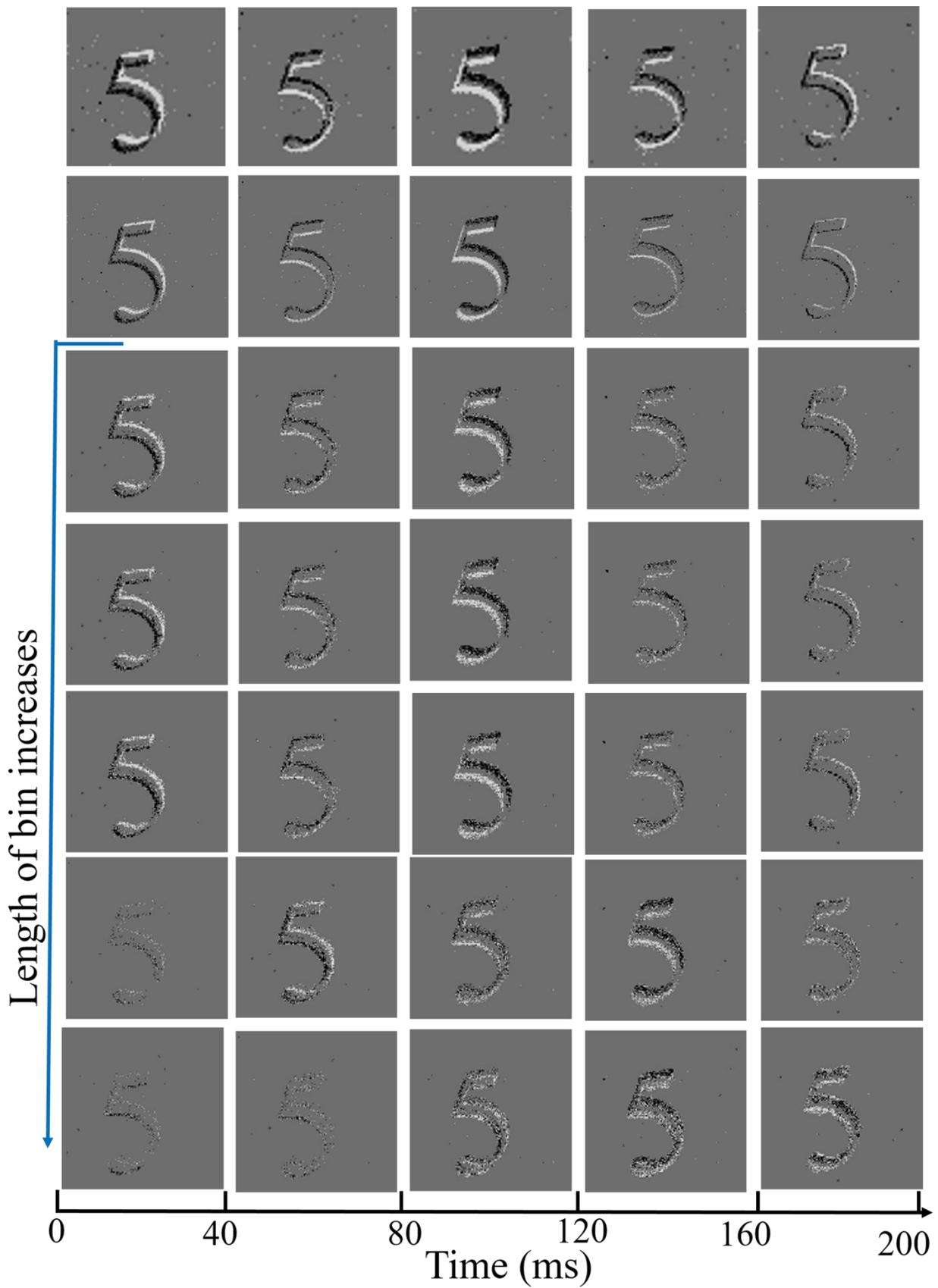

**FIGURE 7** | Influence of the length of the bins in the rate function of point process. From top to bottom are the reconstructed frames of LR event-stream image, groundtruth, SR event-stream images with bin of 20 us, 1000

us, 10 ms, 20 ms, 80 ms. From left to right are the time line. As the DVS recording is in the time window of 200 ms, the reconstructed frames are constructed every 40 ms, to show the temporal dynamics of the event streams.

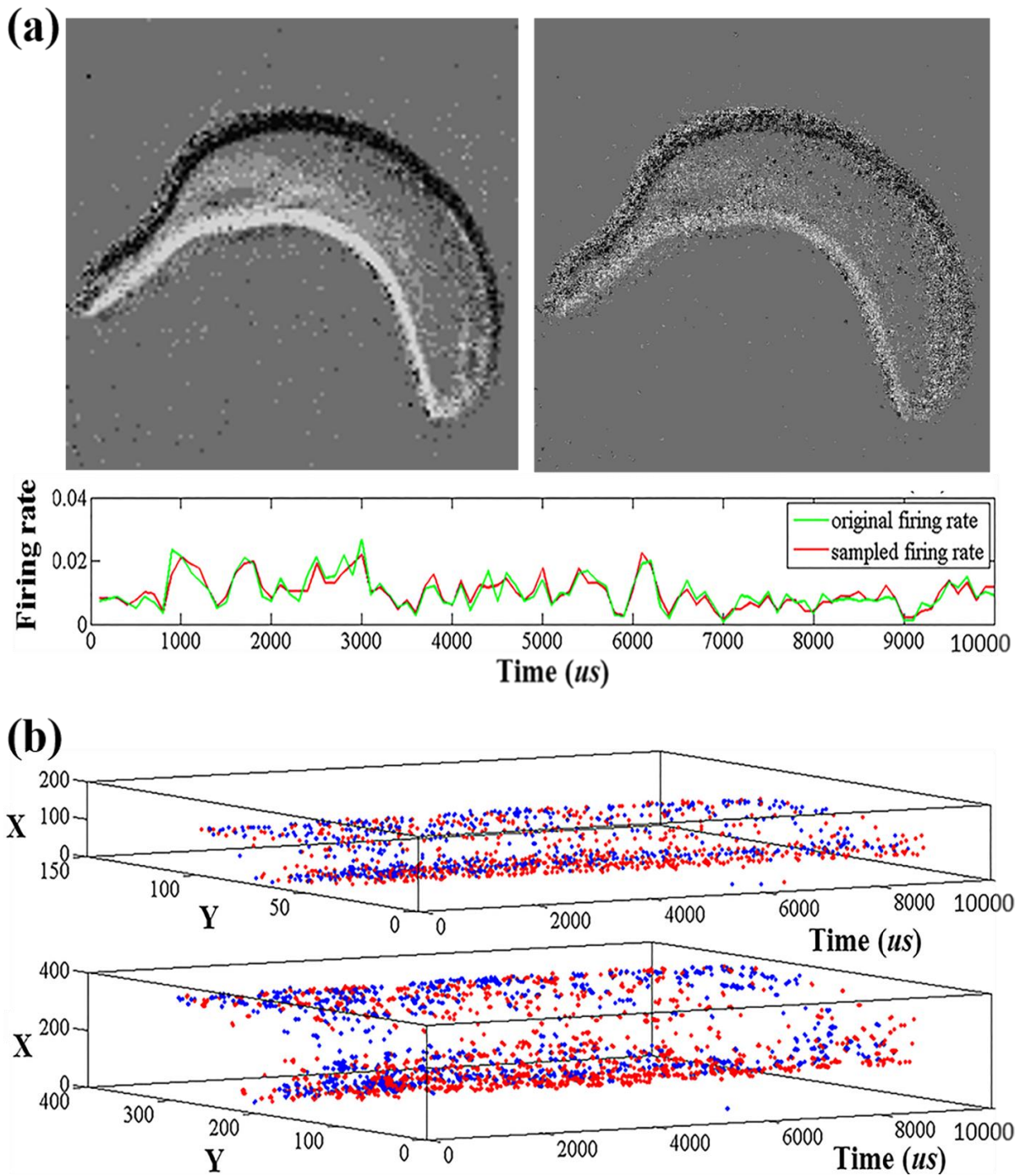

**FIGURE 8|** Upscaling of a spatiotemporal event streams of a banana by a factor of 3. **(a) Top**: Reconstructed frames of LR (left) and HR (right) event streams. Both are in 384×384 size. **Bottom**: total rate function curves of the original LR DVS input (green) and upscaled HR event streams (red). **(b)** Three-dimensional spatiotemporal ON events (red) and OFF events (blue) of LR (top) and HR (bottom) event streams.

**(a)**

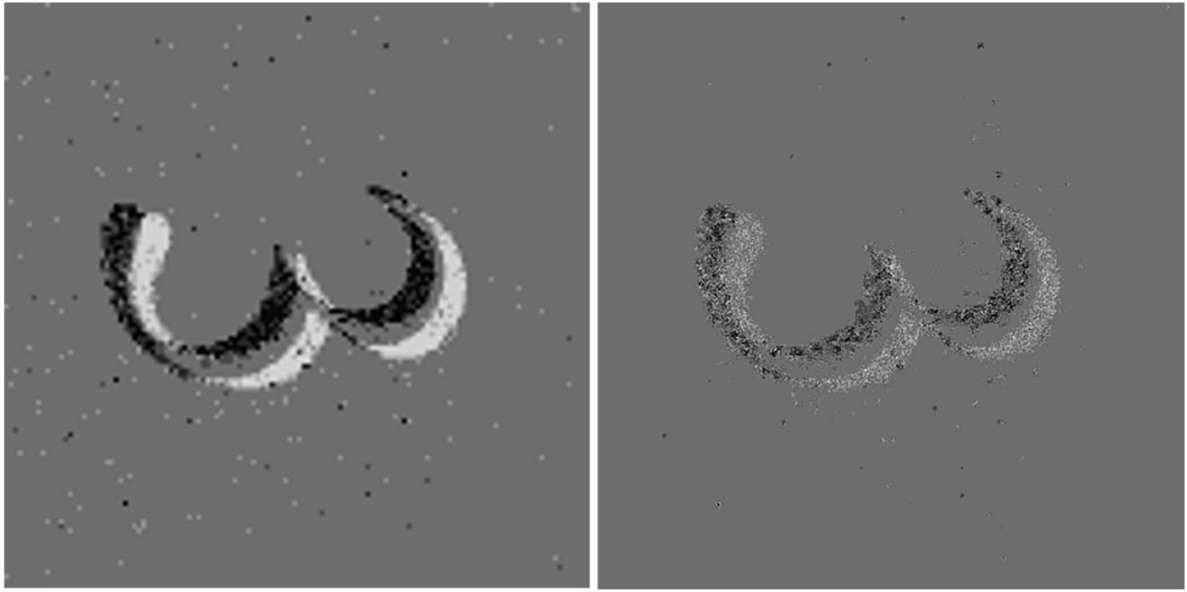

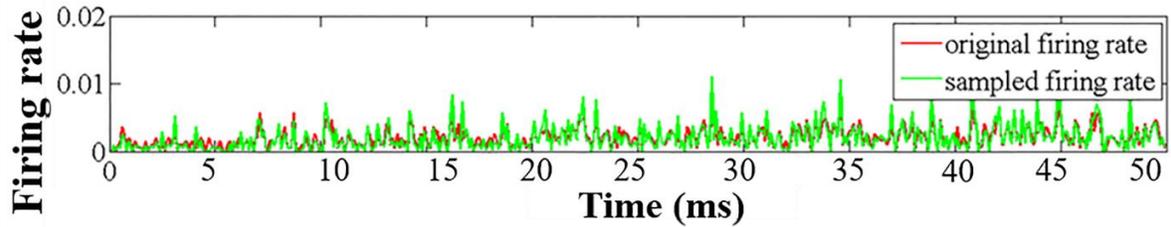

**(b)**

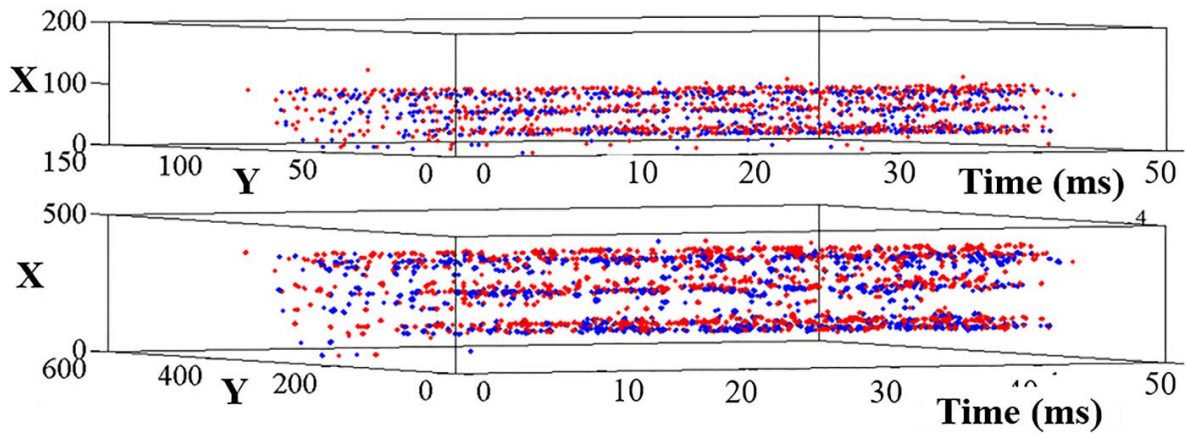

**FIGURE 9 |** 4×upscaling of the spatiotemporal event stream of a digit 3. **(a) Top**: Reconstructed frames of LR (**Left**) and HR (**Right**) event streams in an identical size of 512×512. **Bottom**: total rate function curves of the original LR DVS recording (green) and HR event streams (red). **(b)** Three-dimensional spatiotemporal ON events (red) and OFF events (blue) of LR (top) and HR (bottom) event streams.

# 4    Discussions

To overcome the limitation of physical limit of the DVS sensor such as large pixels and relatively small fill factors, this paper proposes a two-stage SR scheme to obtain a HR spatiotemporal event streams from a LR DVS recording. Different from the two-dimensional intensity reconstruction from the event-based input reported in (Bardow, Davison, and Leutenegger 2016, Reinbacher, Graber, and Pock 2016) which loses the temporal information, our method aims to generate HR three-dimensional spatiotemporal event streams. Different from the SR of the two-dimensional frame-based image, the DVS recording we processed in the proposed method is three-dimensional spatiotemporal event stream. The frame-based SR schemes cannot be directly transferred in the spatiotemporal event streams. To handle the event sequence of each pixel, the Poisson point process is used to model the event sequence. Each point process can be described as a combination of a rate function and the number of point-events. Then by a resolution enhancement method, the event number of the expected HR event-stream image is generated. The rate function of each event sequence is represented as the normalized PSTH. PSTH of each pixel of the LR DVS input is easily calculated. The rate function of each HR pixel can be generated from the LR event streams. For the DVS recording, a single event is not crucial for the formation and visualization of the visual pattern. It is the spatiotemporal cluster of the events that form the shape of the object. The proposed method aims to generate the HR event streams with high spatial texture detail and the same temporal properties with the DVS input.

The proposed two-stage SR method in this work is flexible and effective. In the proposed SR method, the event number of each new pixel and the rate function of each new pixel are firstly to be generated. Improvement of the method for generating the event-count map of the expected HR event streams can be realized with a more complex models. Our method is effective to recover the visually high texture detail and has achieved good performance under two different magnification factors. The proposed method has many potential applications in many dynamic vision processing systems.

In our method, the generation of temporal events is addressed through simulating a one-dimensional nonhomogeneous Poisson process. As is known, two simulated Poisson processes of the same rate function are nearly impossible to have the same timing information although they have the same rate function across time line. Given the event number and rate function of each pixel, temporal events can be sampled by the simple and efficient thinning algorithm or other effective event sampling methods. In addition, each event in the DVS image is a three-dimensional signal containing two-dimensional spatial information and a time information. As such, each DVS recording can be considered as a three-dimension spatiotemporal point process which is a multivariate point process. Then our method can be extended with the use of simulation of multivariate nonhomogeneous Poisson process (Saltzman et al. 2012) which may lead to significant improvement of SR performance.

The generation of event number of each new pixel is addressed based on the resolution enhancement of the event-count map. Example-based methods which have shown better performance in resolution enhancement may be more effective for upscaling the event-count map by recovering more details of the spatial textures. Recently, deep learning based methods (Dong et al. 2014, Wang et al. 2015) are able to learn the end-to-end map between HR and LR event-count maps. The main disadvantage of this kind of methods is the shortage of DVS recordings. If enough DVS recordings are available, deep learning based method may be more effective. While the interpolation method, e.g. nearest-neighboring, linear, or bicubic filtering, can be very fast, they oversimplify the SR problem and usually yield solutions with smooth textures. In addition, the firing rate is calculated using a spatiotemporal filter on the local neighbor pixels. In our experiment, the filter is reduced to a simple 3 × 3 matrix. A more complicated spatiotemporal filter may be helpful for preserving the time property of the events.

## 5    Acknowledgments

This work was supported by the Project of NSFC No. 61327902；SuZhou-Tsinghua innovation leading program 2016SZ0102.

**TABLE 1| Statistics of RMSE results by repeating each SR reconstruction 10 times**

| DVS recording | max RMSE | Min RMSE | Mean RMSE | Standard deviation RMSE (10-4) |
|---|---|---|---|---|
| banana | 0.0798 | 0.0795 | 0.0797 | 1.014 |
| hand | 0.0784 | 0.0779 | 0.0782 | 1.651 |
| Digit 5 | 0.0565 | 0.0561 | 0.0564 | 1.448 |
| Wooden fish | 0.0980 | 0.0976 | 0.0978 | 1.181 |
| cup | 0.1011 | 0.1009 | 0.1010 | 9.391 |
| Digit 2 | 0.0571 | 0.0567 | 0.0569 | 1.180 |
| Scene of digits | 0.0525 | 0.0522 | 0.0523 | 1.147 |
| Digit 3 | 0.0442 | 0.0437 | 0.0439 | 1.475 |
| fork | 0.0658 | 0.0653 | 0.0656 | 1.544 |

**TABLE 2|  DFRF results of SR with different magnification factors**

| DVS recording | ×3 magnification | ×4 magnification |
|---|---|---|
| banana | 1.14% | 2.46% |
| hand | 1.14% | 1.62% |
| Digit 5 | 1.95% | 2.21% |
| Wooden fish | 1.67% | 1.95% |
| cup | 1.22% | 1.82% |
| Digit 2 | 2.31% | 2.38% |
| Scene of digits | 2.07% | 2.68% |
| Digit 3 | 3.10% | 5.94% |
| fork | 1.85% | 2.79% |